# Planning with Brain-inspired AI

Naoya Arakawa

The Whole Brain Architecture Initiative

naoya.arakawa@wba-initiative.org

## Abstract

This article surveys engineering and neuroscientific models of planning as a cognitive function, which is regarded as a typical function of *fluid intelligence* in the discussion of general intelligence. It aims to present existing planning models as references for realizing the planning function in brain-inspired AI or artificial general intelligence (AGI). It also proposes themes for the research and development of brain-inspired AI from the viewpoint of tasks and architecture.

## Introduction

This article surveys engineering and neuroscientific models of planning as a cognitive function. It aims to present existing planning models as references for realizing the planning function in brain-inspired AI or artificial general intelligence (AGI). It also proposes themes for the research and development of brain-inspired AI from the viewpoint of tasks and architecture. This survey may develop into specific requests for research. Planning is an important cognitive function for general intelligence in that it solves problems on the fly without new learning. In psychological discussions of general intelligence, planning is regarded as a typical function of *fluid intelligence*.

In engineering, planning has been studied since the dawn of AI. While there are solutions for symbolically formulated tasks, solving problems in the real world is still difficult. Regarding non-human animals, Caledonian crows, for example, are known to solve planning tasks, and other animals may plan in order to survive in the wild. In the mammalian brain, the prefrontal cortex (PFC) is known to be involved in planning.

In the following, planning is outlined, and then engineering and neuroscientific models are introduced. The article then considers a way for creating brain-inspired models for planning with regard to tasks, baseline systems for evaluation, and whole brain architecture.

## What is planning?

In this article, planning is defined as a task of generating an action sequence (representation) that satisfies constraints of the given representation of the goal and of the current situation. In planning, a representation of an action sequence is generated "in the mind" before moving to action, rather than solving the problem by trial and error in the real world.

In planning, a search for a route to the goal is required. Such a process is also called means–ends analysis, in which the route contains a series of subgoals. In the real world, plans may have to be altered as the external situation changes.

# Components of planning

The following components are required for planning.

## Internal Representation

The following are representations related to planning:

- Representation of the current situation

## Representation of goals

- Representation of applicable actions
- Representation of action results

According to Fuster [1](p.200), these representations, except for perceivable parts of the representation of the current situation, are not perceivable; they are imagined, recalled or predicted. Obtaining (imagining) the representation of changing situations, including the result of action, can be regarded as (mental) simulation.

Actions do not have to be explicitly represented. For example, when the result or goal of action is represented, action could be automatically executed by a policy without its representation being formed.

## Evaluation

Evaluation of the result of action is used to prioritize partial paths in means–ends analysis, and to determine which overall plan should be executed.

## Search

Planning requires a search to find a path to the goal. The following cognitive functions are considered to be involved in a search.

- Backtracking

  This refers to searching unsearched paths (or options that have not been tried) when a path does not work or is unlikely to work. Backtracking requires the agent to memorize searched routes (or tried options).

- Search strategy (heuristics)

  The search is performed on a network (or tree structure) that branches out depending on which policy is selected. Breadth-first or depth-first searches are used as a search strategy, and "pruning" is performed to discard non-promising paths.

- Maximizing the objective function

    In planning, it may be required to simply find a route that achieves the goal, or to find the route that maximizes a value, such as reward.

### Working memory

Working memory is one of the cognitive functions required for planning or a route search [2]. For example, backtracking requires the agent to memorize searched routes (for the short term or medium term).

Working memory has been defined in various ways since the terminology was introduced by Baddeley and Hitch in 1974 [3]. While working memory, in the narrow sense, is the memory of the most recent past that can be examined in delayed memory tasks, we use the term more generally to refer to the memory of any objects of mental operation. Thus, working memory is discussed in terms of the most recent past memory and the prospective memory, which is the representation of things to be realized in the future by planning. Fuster refers to the former as *working memory* and to the latter as *the preparatory set* [1](p.7).

### Attention and inhibition

Since planning is a time-consuming task, it requires concentration on the task over a certain period of time by maintaining attention, or the ability to return to the middle of a task after interruption. Inhibition here means **not** to move attention (not to be distracted). Meanwhile, if the agent does not perform task switching, it becomes a case of *persistence* (lack of *cognitive flexibility*). Thus, proper executive control is performed with (or is the same as) proper attention and inhibition.

# Planning tasks

A typical planning task involves the performance of more than one action, while information on the goal, actions that can be used, and the results for actions are given in advance, and an action series is executed to perform the task without trial and error after the execution starts.

The following are examples of planning tasks:

## Tower of London (Tower of Hanoi)

Tower of London/Hanoi is often used as a planning task to test the human cognitive function [1](p.201). While it is considered suitable for testing the function itself, excluding spatial cues [2], it is not a proper planning task in that there is a solution that does not involve a search. In real life, a clean-up can be considered a similar task.

## Route search

The task is to find the (shortest) route to reach the goal from the starting point.[1] If the agent does not solve it by trial and error in the real environment, it is a planning task. The prohibition of trial and error can be enforced by time restrictions or the prohibition of multiple passes on a path (cf. one-stroke writing restrictions [4] and the *Porteus Maze test*). It is also possible to place restrictions on the order by setting a precondition that a certain action should be followed via the passage of a specific route.

## Tool use

Caledonian crows are known to perform planning-like behavior with tools [5].[2] Specifically, they assemble series of actions by using tree branches or stones as tools to obtain another tool and finally to obtain food. In these situations, they use tools as an extension of the body or learned physical laws. Though these behaviors are interesting, they cannot be achieved with planning alone. A task to reach the goal by building bridges with sticks is another example of route search tasks with tools.

## Other relevant tasks

In the N-back[3] and ordered object (self-ordered) tasks[4], it is necessary to memorize multiple objects that appeared in the most recent past. Such working memory is called *monitoring*, and it is similar to prospective memory in that it involves multiple objects.

In research examining the PFC, task-switching tests, such as the *Wisconsin Card Sorting Test*, are often employed to examine *cognitive flexibility* and *perseveration*. While it is not apparent how task switching is related to planning, common sub-functions such as working memory, attention, and inhibition may be involved. Note that inhibition is regarded as important in *delayed gratification tests* (for example, the *Marshmallow Experiment*), which examine relatively long-term behavior planning in which priority is given to a larger reward obtained after the currently available one.

---

[1] For examples abstracting route search, visit: https://www.geeksforgeeks.org/backtracking-algorithms/
[2] For more academic discussions, refer to the publications of Alexander Taylor: https://unidirectory.auckland.ac.nz/profile/alexander-taylor
[3] https://en.wikipedia.org/wiki/N-back
[4] See §7.4 of *Handbook of Frontal Lobe Assessment*, OUP Oxford (2015)

# Engineering models

## Symbolic models

Here, symbolic models refer to information processing models that are not analog/distributed (like neural networks), and instead use only symbolic operations. The General Problem Solver (GPS) is a classic symbolic model for planning. Problems given to GPS are formulated with an initial state and a goal state, differences between states, and operators to fill the differences. A state from which a goal can be reached by applying an operator is called a subgoal. GPS repeatedly (recursively) obtains subgoals, and finds a series of operators from the initial state to the goal state (this is also called means–ends analysis). STRIPS, developed in the 1970s, is an example of GPS. Note that there is a dynamic programming algorithm, such as Dijkstra's algorithm, when an optimal solution is obtained for a route search (planning) in which the cost of local routes is known. Symbolic models have practical problems in that they require formalized specifications in advance and a large amount of time for exhaustive searches in large problem spaces, and the probability may not be handled well. Note that there are hybrid systems that use neural networks to symbolize the environment and solve the planning problem with a symbolic model (e.g., see [6]).

## Learning models

In symbolic models, knowledge has to be given in advance, whereas learning models acquire knowledge by themselves. The knowledge representation of learning models, in most cases, is analog and distributed, and it could also represent probabilistic knowledge.

### Reinforcement learning

Reinforcement learning (RL) is a type of machine learning in which action policies are learned so that a series of actions the agent takes yield the largest reward [7]. In the mammalian brain, the basal ganglia (BG) or loops consisting of the frontal lobe cortices, BG, and thalamus[5] (*FLC–BG–thalamic loop*) are thought to perform reinforcement learning.

Model-based reinforcement learning uses a model for estimating the environment. In particular, an RL system that uses a model at the time of action can perform planning as defined in this article. The model probabilistically predicts inputs or assumed states from past inputs (including sensory inputs, agent actions, and rewards), with which the system determines action. Models are preferably to be learned rather than hand-coded. Since the brain can predict its next states and reward (in PFC in particular) or it has the model of the environment, it is thought to be performing model-based RL.

---

[5] See https://grey.colorado.edu/CompCogNeuro/index.php/CCNBook/Motor .

The Monte Carlo Tree Search (MCTS) method is widely used for model-based reinforcement learning performing action-time prediction. For example, *MuZero* [8] has learned the model (environment) and achieved good results in games requiring look-ahead, such as Go, and also in video games.

RGoal Architecture [9] applies the hierarchical structure in *hierarchical reinforcement learning* to plan as in model-based RL.

### Gated RNNs

In LSTM (Long Short-Term Memory) and GRU (Gated Recurrent Unit), which are kinds of RNN, "gating" determines whether its states are maintained or terminated. Planning/route search tasks require working memory to keep track of subgoals currently pursued and subtasks already examined. It should be empirically examined whether gated RNNs can learn planning tasks as they are, or by adding certain mechanisms.

### Memory networks: algorithm learning by training RNN

It is known that RNN (LSTM) can learn algorithmic behavior by supervised learning or reinforcement learning. Such mechanisms are called memory networks [10]. Since planning also is implemented in certain algorithms, *memory networks* may be able to learn them. In fact, a memory network called *differentiable neural computers* (DNC) learns route planning.

### Attention models and MERLIN

In recent natural language processing, systems using "attention models" instead of RNN have been performing well [11][12]. Attention models use context-dependent associative memory.

MERLIN [13] is a system that performs well in various tasks, including maze tasks, by combining gated RNN, attention models, and reinforcement learning. As MERLIN can memorize its internal states at each point in time, it can use knowledge of a situation encountered only once in the past.[6] It is reported that MERLIN showed "hierarchical goal-directed behaviour" even though it was not explicitly programmed, suggesting that it has a planning function.

# The functions of PFC in planning

This section gives a summary of the often-highlighted involvement of PFC in planning. It mainly refers to the work of Passingham et al. [15], Fuster [1], and Tanji et al. [16] for the functions of PFC. Passingham et al. suggest that the function of PFC is to "generate goals that are appropriate to the current context and current needs" [15](p.220). Fuster writes that the basic function of PFC is "the representation and execution of new forms of organized goal-directed action" [1](p.1), and points out the importance of temporal integration in the executive function, including planning (ibid. pp.7 & 201). Tanji et al. review the executive function of the lateral PFC from the viewpoint that the central function of the PFC is integrative planning [16].

---

[6] Similar ideas are explored in [14].

Hereafter, brain parts related to the cognitive functions necessary for planning (see above) are listed. It is often assumed that action selection (decision-making) and execution are performed with the FLC–BG–thalamic loop.

## Representation

This section addresses a particular question: Which parts of the PFC are related to the representations mentioned in the section on the components of planning?

- Representation of the current situation

    The representation of the current situation is thought to be handled in parts that process sensory information. For vision, the dorsal (where) path responsible for information processing of motion and location sends its output to BA8,[7] including the frontal eye field (FEF) and dorsolateral PFC. The ventral (what) path, which is involved in the identification of (visual) objects, sends information to the ventrolateral PFC (BA45 and BA47) [15]. The orbital PFC also receives visual information from the ventral path.

- Representation of goals

    The representation of goals is defined here as a "perceptual image of a situation that is not necessarily present but to be realized." Passingham et al. claim that the function of the (lateral) PFC is the generation of goals [15](Chapter 6). Yamagata et al. claim that the dorsal PFC and premotor cortex maintain goal representation [17].

- Representation of applicable actions (policies)

    Passingham et al. claim that the medial PFC "contributes to evaluating and choosing among actions based on associations with outcomes, in relation to current needs" [15](Chapter 3). The medial PFC has bidirectional connections with the hippocampus and may receive the sequential representation of the result of simulated action from the hippocampus. In the frontal lobe, the closer an area is to the motor area, the more direct its involvement in movement, and the more forward it is, the more preparatory its behavior [1](p.320).

---

[7] BA stands for Brodmann Area.

## Evaluation

In reinforcement learning, policies are evaluated with regard to reward. Passingham et al. assume that stimulus evaluation takes place in the orbital PFC, with action evaluation in the medial PFC, where situations and policies are supposed to be respectively evaluated. Evaluative functions in the medial PFC vary depending on the areas, and various experimental results have been reported [15].

## Search

Saito et al. [4] discovered that the dorsolateral PFC is involved in the final goal representation of tasks. As a search is a typical planning task, the other cognitive functions mentioned in this article should be integrally involved.

## Working memory

Working memory is divided into the retrospective memory (of the recent past) and the prospective memory (the representation of things that should be realized in the future; see above).

In both cases, the way in which (short-term) memory is realized in the brain is an issue. Short-term memory can be realized by recurrent neural circuits, which may be local ones within the same cortical region [18] or global ones that span multiple cortical regions [1] (pp.270–283). The PFC may be controlling the retention of memory by attention control (below) rather than its being the memory itself [19][1](p.394).

As for prospective memory, works by, for example, Passingham et al. [15], Tanji et al. [16], and Owen [2] highlight the function of the dorsolateral PFC.

## Attention and inhibition

According to Fuster [1], "attention consists of two antagonistic and complementary cognitive processes, one of focusing, inclusive, and the other inhibitory, exclusionary" (p.142). The former involves the dorsolateral PFC, and the latter involves the ventromedial PFC (p. 254). He identifies the former as the preparatory set (or prospective memory; see above) for preparing action (p.143).

Fuster [1](p.319) also points out that the anterior cingulate cortex (ACC or the medial PFC) is active in preparing tasks and actions that require attention (e.g., Stroop tasks) [20][21][22], and that the ACC and the inferior PFC (or ventrolateral PFC) send inhibitory signals to the thalamus.

# Architecture

The brain regions introduced above (excluding the basal ganglia and thalamus), and the main connections between them, are shown in Fig.1.

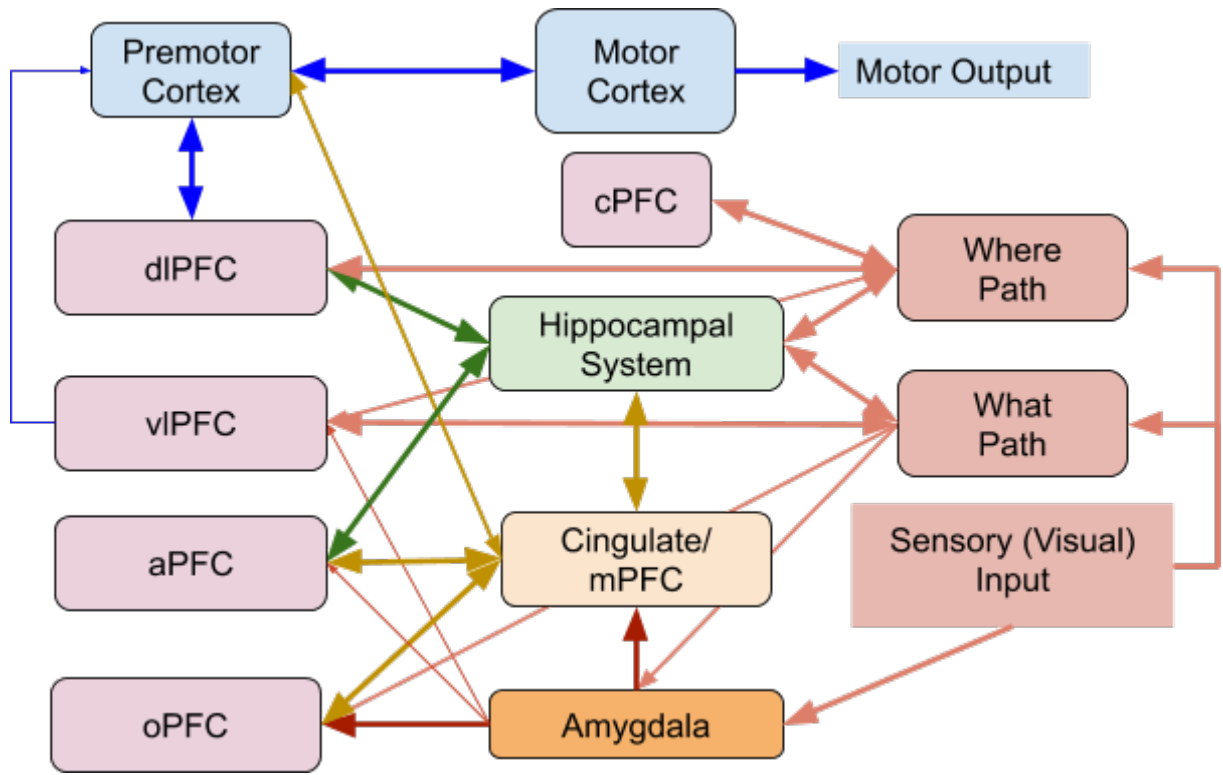

Fig. 1

Unless otherwise noted, connections are from the work by Passingham et al. [15]. The FLC–BG–thalamus loop is associated with all the cortical areas in the figure, and is involved in controlling with reinforcement learning. In the following, the parts are briefly described.

**Where Path**: Coding the locations (coordinates) and movement of objects in the environment.
**What Path**: Coding what is in the environment.
**cPFC** (Caudal PFC): Including FEF and involved in visual attention. With output to areas related to eye movement such as the superior colliculus (not shown in the figure).
**Premotor Cortex** and **Motor Cortex**: Involved in motor execution
**dlPFC** (Dorsolateral PFC) (BA8, BA9, BA10, BA46): Connections mainly with the dorsal (where) path and hippocampal areas. As mentioned above, it is assumed to be involved in prospective memory, attention, and setting final goals.
**vlPFC** (Ventrolateral PFC) (BA45, BA47, BA44): Connections mainly with the ventral (what) path. The connections shown with the dorsal path in Fig.1 are taken from the work by Tanji et al. [2008]. The vlPFC is assumed to be involved in setting perceptual goals.

**oPFC** (Orbital PFC): Receiving affect information from the amygdala and perceptual information from the ventral (what) path. Passingham et al. [15] assume that the area evaluates and selects perceptual objects.

**aPFC** (anterior/polar PFC): Mainly BA10. See [23] for connections and [24] for functional hypotheses. It is assumed to be involved in multitasking (switching tasks).

**Cingulate/mPFC** (Medial PFC): Consisting of agranular cortices (ACC BA24, anterior limbic cortex BA32, and inferior limbic cortex BA25) as prefrontal areas. While connections and functions differ according to areas, they are generally related to reward (evaluation).

**Hippocampal System**: Consisting of several areas centered on the hippocampus. It integrates the where and what information to create the representation of situations (episodes), and to fix them in long-term memory (for its relation with PFC, see [25]).

**Amygdala**: It is assumed to receive perceptual and affective information and to be involved in the (affective) evaluation of situations [26]. There are connections with the hippocampus, but these were omitted from the figure.

BA44 and BA45 in the vlPFC have linguistic functions connected with the superior temporal gyrus related to hearing. The dlPFC and aPFC have connections with the multimodal area in the same gyrus.

Fig.1 suggests that there are few strong connections between PFC areas. If two of these areas are to communicate, they will do so via another area to which they are both connected.

The planning model using the area shown in Fig. 1 will be discussed again in the section "A Proposal – Whole brain architecture" below.

# Computational neuroscientific models

This section introduces a model of working memory with the FLC–BG–thalamic loop and a PFC model that solves the Tower of Hanoi task. Neither solves the planning task per se.

## PBWM

The PBWM (prefrontal cortex basal ganglia working memory) model is a model of the executive function in the brain [27], and it is summarized as follows:
- It controls whether internal states are kept or released.
- The internal states are maintained in the PFC.
- Control is achieved through reinforcement learning in the FLC–BG–thalamic loop.

The control of whether internal states are kept or released is the "gating" function, as mentioned in the section describing gated RNNs above. It should be empirically verified whether the PBWM can learn any planning task as it is, or if an additional mechanism is required.

## Stewart and Eliasmith's models for the Tower of Hanoi

Stewart and Eliasmith propose a computational neuroscientific model that solves the Tower of Hanoi task [28]. The model is summarized as follows.

- The model does not learn; the algorithm is hand-coded.
- The algorithm is deterministic; it does not involve a search.
- The states of the Tower of Hanoi – the objects of attention, the object of movement, and the target – are symbolically given.
- Models (architecture) have the visual cortex, motor cortex, other neocortices, basal ganglia, and thalamus as modules.
- It uses the Leaky Integrate-and-Fire (LIF) spiking neuron for its neuron model.
- It uses Vector Symbolic Architecture to cope with the binding problem.

As the model uses a hand-coded deterministic algorithm, it would be difficult to extend it to general intelligence.

# A Proposal

This section discusses a way to create a planning model of the mammalian brain. More specifically, it discusses tasks, baseline systems for evaluation, and whole brain architecture for planning.

## Tasks

The "planning task" section looked at the Tower of London/Hanoi and route search tasks. While the Tower of London/Hanoi task is widely used to investigate human planning functions, it has a problem with its deterministic solutions. In route search tasks, agents may rote-memorize solutions after going over specific terrains many times. Thus, route search tasks in which the learning and test environments differ are desirable for testing the planning function. Such environments can be generated with tools.

When preparing a task, one should keep in mind the extent to which information on the environment is given symbolically. In general, the greater the extent to which information is given symbolically, the easier the task. However, for humans and animals, symbolification may not make the task easier, because it reduces ecological (spatial) cues as the level of abstraction increases.

## Baseline

Engineering models are used as the baseline for evaluation. When a task is symbolized, there is no problem for symbolic models dedicated to planning (e.g., STRIPS and other route search algorithms) to solve it. Among the brain-inspired engineering learning models mentioned above, memory networks and MERLIN can be used as the benchmark. PBWM can be a computational neuroscientific benchmark. Other models that may not have structures similar to the brain could still be used as benchmarks.

To evaluate and compare the planning function itself, tasks may be given symbolically. In this, systems would be evaluated with the learning capability of planning and biological plausibility of architecture.

## Whole brain architecture

This section discusses the planning of mammalian brain models, from simpler ones to more complex ones. For brain architecture, see Fig. 1.

Planning or means–ends analysis starts with the representation of the current situation and goal, creates a representation sequence consisting of the representation of subgoals, operators (actions), and action results, and is completed by connecting the representations of the current situation and the goal. In this process, working memory (prospective memory) is required, when options are tried "in the head" to create the representation sequence. Marking search paths already considered is a role of working memory in planning, and identifying its biological mechanism is an issue. If this is done by episodic memory, the involvement of the hippocampus area should be considered. Backtracking tasks in the real world require task switching, which is supposed to involve the anterior PFC. This region may also be involved in backtracking in mentally simulated tasks (i.e., planning). If the executive control (attention and inhibition) of planning is the result of reinforcement learning, then the FLC–BG–thalamic loop (as assumed in the PBWM model) or PFC trained by reinforcement learning would be involved (as assumed by Wang et al. [29] and Yamakawa [30]).

In planning or means–ends analysis, it is also necessary to predict or simulate the result of the action. The location of the simulator in the brain can be in the PFC, the hippocampus, or a combination of both. The replay or preplay of past situations performed in the hippocampus may provide approximate solutions based on past experience or episodes in the task space.

The lateral PFC can focus on sensory information with attention. For notably important pieces of perceptual information, attention is focused through the conscious access or "avalanche of consciousness" involving the PFC and posterior perceptual areas [31].

Episodic memory around the hippocampus seems to be accessed with attention from the dorsolateral PFC (Fig. 1). Recalling subgoals in planning (means–ends analysis) could be done by recalling episodes or the representation of generalized episodes from long-term memory.

In a route search, information processing of "where and what" is required. In the brain, where and what information is processed in different pathways (the dorsal and the ventral paths, respectively). For proper evaluation of biologically inspired planning functions, it would be desirable to use a shared brain-inspired visual information processing mechanism that separately processes where and what information in baseline models.

The following section outlines a planning model inspired by the brain. The algorithm is simple, so that its learning will be easy. It is up to the reader to decide whether it is an appropriate model of the brain.

- Set the overall goal to the prospective memory with regard to the current situation.
    - When the goal is perceivable:
      A module corresponding to the ventrolateral PFC selects a perceivable object as the goal.
    - When the goal is not perceivable:
      Past situations are recalled from the current situation in a module corresponding to the association area around the hippocampus, and a module corresponding to the dorsolateral PFC selects a goal from the representation of the past situations.

- Set subgoals to the prospective memory from the current situation and overall goal. The related brain areas are the same as with the setting of the overall goal. Subgoal recall is weighted by the following scales:
    - Conditional probability of the subgoal, given the features of the current situation and goal.
    - The feature distance relative to the current situation and goal (the nearer a subgoal is to the overall goal, the more it is prioritized).
    - Expected reward (the greater the expected reward, the greater the subgoal's priority).

- Planning by simulation
  Check whether a policy that reaches the other goal from the current or highly prioritized subgoal can be recalled. Increase the priority of the subgoal based on the evaluation of the found policies.

Repeat this until a policy sequence linking the current situation to the overall goal is found.

Policy recall can be done in a module corresponding to the associative areas around the hippocampus or the FLC–BG–thalamic loop.

# Conclusion

Finally, what are the engineering implications of models and the neuroscientific implications discussed here?  While planning in engineering (AI) is important for the engineering realization of general intelligence, if the task is not given symbolically, current technology does not allow it to be solved at the human level, and thus learning from the brain makes pragmatic sense.  Meanwhile, if planning can be performed by architecture that mimics the brain, it will be useful for neuroscience as a model for planning in the brain.

# Acknowledgment

This article was made possible with funding from the Kakenhi project "Brain information dynamics underlying multi-area interconnectivity and parallel processing."  I am also thankful to valuable comments from Yuji Ichisugi, Susumu Ota, Gideon Kowadlo, and Hiroshi Yamakawa, which I received upon writing this article.